\documentclass[11pt]{article}

\usepackage[final]{acl}

\usepackage{times}
\usepackage{latexsym}
\usepackage{bbding}
\usepackage{tabularx}
\usepackage{threeparttable}
\usepackage{pifont}
\usepackage{caption}
\usepackage{booktabs}
\usepackage{multirow}
\usepackage[T1]{fontenc}
\usepackage[utf8]{inputenc}
\usepackage{microtype}
\usepackage{inconsolata}
\usepackage[table]{xcolor}
\usepackage{graphicx}
\usepackage{listings}

\newcommand{\cmark}{\ding{51}}
\newcommand{\xmark}{\ding{55}}

\title{BabelDOC: Better Layout-Preserving PDF Translation via Intermediate Representation}

\author{
  \textbf{Qi Yang\textsuperscript{1,}\thanks{Equal contribution.}},
  \textbf{Xiangyao Ma\textsuperscript{2,}\footnotemark[1]},
  \textbf{Xiao Wang\textsuperscript{1}},
  \textbf{Hao Wang\textsuperscript{1}\thanks{Corresponding author.}},
  \textbf{Rui Wang\textsuperscript{3}}
\\
  \textsuperscript{1}School of Computer Engineering and Science, Shanghai University, Shanghai, China \\
  \textsuperscript{2}Funstory.ai Limited, Hong Kong SAR, China \\
  \textsuperscript{3}Department of Computer Science and Engineering, Shanghai Jiao Tong University, Shanghai, China \\
  \texttt{yang-qi@shu.edu.cn}, \texttt{aw@funstory.ai}\\
  \texttt{\{22122792,wang-hao\}@shu.edu.cn}, \texttt{wangrui12@sjtu.edu.cn}
}

\begin{document}

\maketitle

\begin{abstract}
As global cross-lingual communication intensifies, language barriers in visually rich documents such as PDFs remain a practical bottleneck. Existing document translation pipelines face a tension between linguistic processing and layout preservation: text-oriented Computer-Assisted Translation (CAT) systems often discard structural metadata, while document parsers focus on extraction and do not support faithful re-rendering after translation. We introduce \textbf{BabelDOC}, an Intermediate Representation (IR)-based framework for layout-preserving PDF translation. BabelDOC decouples visual layout metadata from semantic content, enabling document-level translation operations such as terminology extraction, cross-page context handling, glossary-constrained generation, and formula placeholdering. The translated content is then re-anchored to the original layout through an adaptive typesetting engine. Experiments on a curated 200-page benchmark, together with human evaluation and multimodal LLM-as-a-judge evaluation, show that BabelDOC improves layout fidelity, visual aesthetics, and terminology consistency over representative baselines, while maintaining competitive translation precision. The open-source toolkit and its interactive downstream applications are publicly available and have attracted over 8.4K GitHub stars and 17 contributors at the time of writing: \href{https://github.com/funstory-ai/BabelDOC}{https://github.com/funstory-ai/BabelDOC}. A demonstration video is available at \href{https://youtu.be/chwrlApH7a4}{https://youtu.be/chwrlApH7a4}.
\end{abstract}

\section{Introduction}

\begin{table}[t]
\centering
\resizebox{\columnwidth}{!}{%
\begin{tabular}{l >{\columncolor{green!10}}c c c c}
\toprule
\textbf{Feature} & \textbf{BabelDOC} & \textbf{PDFMathTrans.} & \textbf{Doc2X} & \textbf{Google / DeepL} \\
\midrule
Formula processing               & \cmark & \cmark & \cmark  & \xmark \\
Footnote translation             & \cmark & \xmark & \xmark  & \xmark \\
Visual style preservation (PDF)  & \cmark & \xmark & \xmark  & Partial \\
In-figure text translation       & \cmark & \xmark & Partial & \xmark \\
Intermediate representation      & \cmark & \xmark & \xmark  & \xmark \\
Auto contextual terminology      & \cmark & \xmark & \xmark  & \xmark \\
\bottomrule
\end{tabular}%
}
\caption{Comparison of capabilities between BabelDOC and representative document translation systems. PDFMathTrans. denotes PDFMathTranslate \cite{ouyang-etal-2025-pdfmathtranslate}.}
\label{tab:quick-comparison}
\vspace{-1.0em}
\end{table}

PDF is extensively utilized across diverse domains---including scientific research, commerce, law, and technical exchange---owing to its cross-platform compatibility and strict visual integrity. Nevertheless, the dominance of the English language in global knowledge production creates significant inequality \cite{hwang2005inferior, ulrichammon2012linguistic}. Language barriers severely impede the dissemination and reuse of visually rich documents, often excluding non-English-speaking researchers from participating fully in the international scientific community \cite{ramirez2020disadvantages, bahji2023exclusion, montgomery2013does}. Achieving high-quality technical translation \cite{schubert2012technical} for PDFs is therefore of practical significance for open science \cite{unescoopen}; however, it remains challenging because PDFs are fundamentally designed for display rather than semantic editing.

Currently, document translation is caught in a technical dilemma. On one hand, traditional Machine Translation (MT) and Computer-Assisted Translation (CAT) tools \cite{johnson2017googles} focus primarily on textual streams, often destroying rich structural metadata during extraction. On the other hand, recent advancements in document parsers (e.g., Doc2X and MinerU) excel at structural extraction but perform mainly \textit{unidirectional} conversion, such as PDF-to-Markdown. They usually discard or abstract away the typesetting metadata required to reverse the process, rendering them insufficient for high-fidelity PDF reconstruction after translation. Furthermore, monolithic end-to-end translation attempts, such as our early precursor PDFMathTranslate \cite{ouyang-etal-2025-pdfmathtranslate}, act as comparatively opaque pipelines. This limits scalability and makes document-level NLP interventions difficult, especially for terminology consistency and cross-page context awareness.

To overcome these bottlenecks, we propose \textbf{BabelDOC}, a translation framework built upon an \textbf{Intermediate Representation (IR)} paradigm. Instead of treating document translation as a black-box end-to-end task, BabelDOC explicitly decouples textual semantics from visual layout information. Unlike unidirectional parsers, our bidirectional IR allows semantic content to be manipulated for advanced NLP tasks while preserving structural metadata for layout-aware PDF reconstruction. The system first parses the source document into a structured IR, isolating elements such as paragraphs, formulas, and graphic bounding boxes. Complex linguistic processing---including translation powered by Large Language Models \cite{brown2020language, dalayli-2023-use}, automated glossary extraction, and cross-column contextual modeling---is executed at the semantic layer. Finally, an adaptive typesetting engine re-maps the translated text back into the original structural constraints, adjusting localized scaling factors to improve visual reconstruction.

Table \ref{tab:quick-comparison} summarizes how BabelDOC differs from representative PDF translation and document parsing systems. By transforming the translation workflow into a transparent, plugin-based architecture, BabelDOC serves as a bridge between the Document Understanding and Natural Language Processing communities. The major contributions of this system demonstration are threefold:
\begin{itemize}
    \item \textbf{An IR-based Translation Paradigm:} We propose a decoupled architecture that translates text at the semantic level while preserving structural layout metadata, reducing the conflict between linguistic processing and visual formatting.
    
    \item \textbf{Document-Level NLP Integration:} Unlike generic parsers, our IR framework supports translation interventions including automatic terminology extraction, customized glossary constraints, cross-page context handling, and mathematical placeholders.
    
    \item \textbf{An Open-Source, Extensible Ecosystem:} Recognizing the challenges and benefits of open-source software in product development \cite{stol2010challenges}, we open-source the BabelDOC backend engine alongside interactive user-facing interfaces. The framework's modular design allows researchers to hot-swap LLMs and OCR engines, fostering further work in cross-lingual document intelligence.
\end{itemize}

A more comprehensive capability comparison between BabelDOC and existing systems is presented in Table \ref{tab:full-tool-comparison}. BabelDOC is designed for two audiences: (1) scientific researchers and professionals who need to consume multilingual literature without losing complex layouts (e.g., math-heavy arXiv papers), and (2) NLP developers who can utilize our Python API to build custom cross-lingual document intelligence pipelines. A 2.5-minute screencast demonstrating the interactive features of BabelDOC, alongside the open-source repository and live demo access, is provided in the supplementary materials.

\section{Related Work}

\paragraph{Document Layout Analysis and Parsing.}
The accurate extraction of visual and structural metadata from PDFs is a prerequisite for downstream document intelligence. Recent advancements have significantly improved structural analysis; for instance, models like DocLayout-YOLO \cite{zhao2024doclayoutyolo} and YOLOv10 \cite{wang2024yolov10} have enhanced high-speed bounding box detection, while tools such as Doc2X and MinerU excel at converting complex PDFs into Markdown or \LaTeX~source code. Specialized tools also exist for localized tasks, such as Mathpix for formula extraction and LayoutReader for reading-order determination. However, these systems function primarily as unidirectional extractors. They are not designed to re-render processed semantic content back into its original visual layout, making them insufficient for end-to-end layout-preserving document translation tasks.

\paragraph{Machine Translation for Visually Rich Documents.}
Machine translation has evolved from statistical approaches \cite{brown1990statistical} to highly sophisticated Neural Machine Translation (NMT) architectures based on the Transformer \cite{vaswani2017attention} and subword tokenization \cite{sennrich2016neural, zhu2020incorporating}. More recently, Large Language Models (LLMs) have demonstrated strong zero-shot translation capabilities \cite{brown2020language}, which can be further enhanced through strategic role-play prompting \cite{shanahan2023role, kong2024better}. Despite these linguistic advances, generic LLMs operate mainly on one-dimensional text streams and do not directly model the layout constraints of PDFs.

To bridge the modality gap, our prior work, PDFMathTranslate \cite{ouyang-etal-2025-pdfmathtranslate}, pioneered an end-to-end pipeline capable of synchronized translation and layout preservation, attracting significant open-source community contributions \cite{ws0516822025, Wybxc2025}. While it demonstrated the feasibility of visual fidelity, its monolithic architecture lacked an explicit intermediate representation. Consequently, it had restricted scalability for semantic processing and structural manipulation, including document-level terminology extraction, cross-column contextual awareness, and complex nested structure rendering. BabelDOC addresses these limitations by introducing a decoupled intermediate representation paradigm.

\section{System Architecture and Methodology}

\begin{figure}[t]
\centering
    \includegraphics[width=0.98\linewidth]{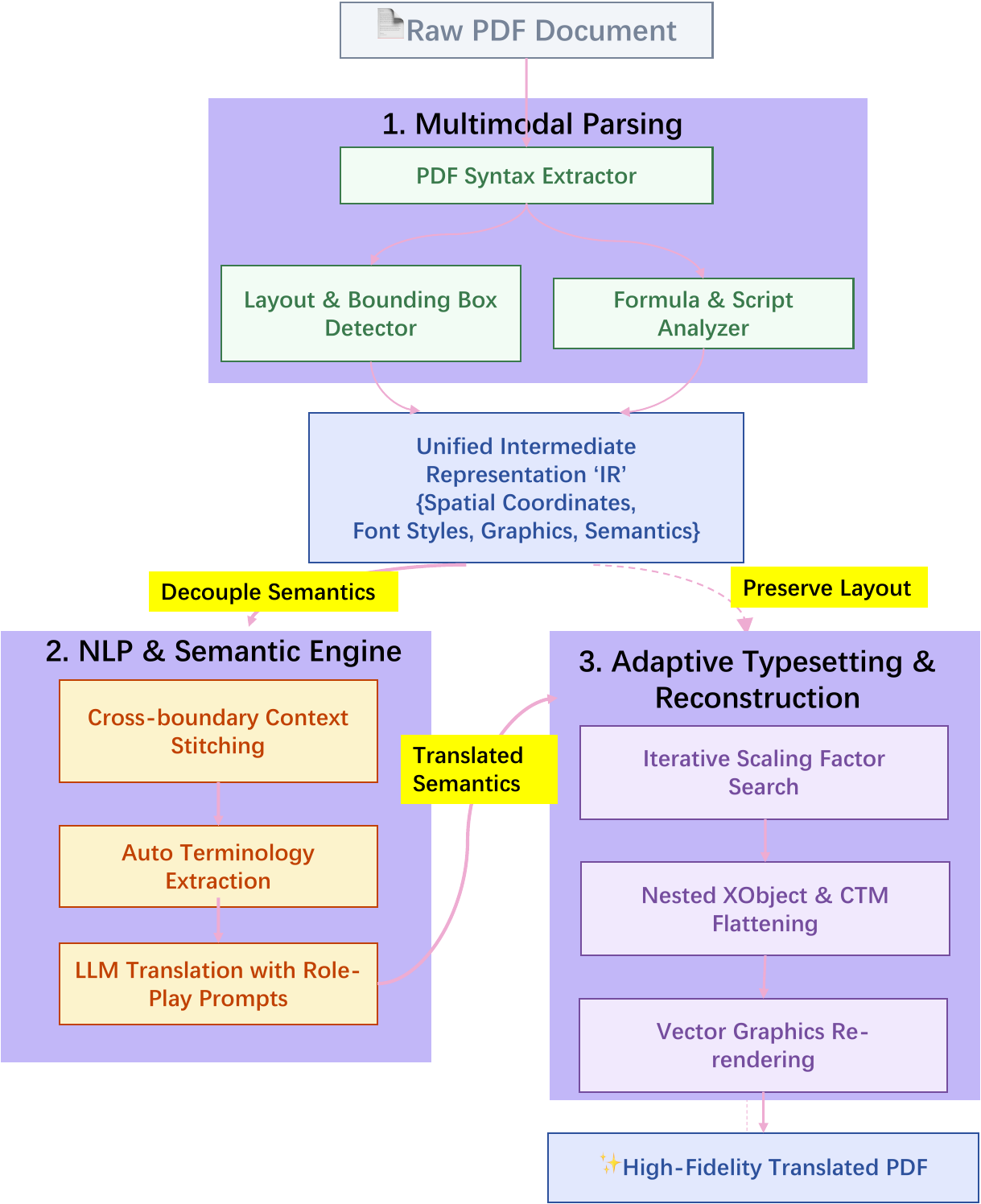}
    \caption{The system architecture of BabelDOC. Raw PDFs are decoupled into a unified intermediate representation, allowing document-level NLP translation interventions to occur independently before the content is adaptively reconstructed via the typesetting engine.}
    \label{fig:arch}
\end{figure}

As illustrated in Figure \ref{fig:arch}, BabelDOC departs from traditional black-box translation pipelines by employing a modular, Intermediate Representation (IR)-centric architecture. By explicitly decoupling visual layout from textual semantics, the system executes linguistic processing at the semantic layer before re-anchoring the output to the structural layout. The workflow is orchestrated through five integrated modules.

\subsection{Decoupled Intermediate Representation}

Since PDF utilizes a specialized imperative syntax for document rendering---ensuring precise visual reproduction across various devices---the format is inherently difficult to parse or manipulate directly for semantic tasks. To facilitate efficient document translation, our system first standardizes the input PDF and parses it to generate a unified Intermediate Representation (IR).

The IR serves as a structured data interface that encapsulates per-page elements, including characters, text lines, graphic blocks, and inline images. Each element is stored with spatial coordinates, bounding boxes, and stylistic attributes, providing a high-fidelity approximation of the document's original content. By preserving raw layout metadata from the source file in a structured format, the IR provides a bidirectional foundation for downstream modules to perform paragraph recognition, translation, and typesetting.

\subsection{Formula and Multimodal Processing}

Mathematical notation poses a significant challenge in scholarly document translation, and its accurate handling without triggering LLM-induced corruption is important. Operating on the IR database, our formula processing module comprises three core components:
(1) \textbf{a script detection unit}, which generates script identifiers (e.g., for superscripts and subscripts) based on the font-size variance of adjacent characters retrieved from the IR;
(2) \textbf{an offset calculation unit}, which is coupled with the detection unit to generate fragment offsets based on established baseline coordinates; and
(3) \textbf{a vector reconstruction unit}, which utilizes these offsets to produce a \textit{vector formula analyzer}. This mechanism allows mathematical notations and other non-translatable structured fragments to be masked as placeholders during the NLP translation phase and reconstructed afterward with high fidelity.

\subsection{Semantic Engine and Contextual Translation}

With structural metadata stored in the IR, translation is elevated from isolated sentence processing to document-level contextual reasoning. Leveraging modern LLMs enhanced by role-play prompting, BabelDOC implements several NLP features at the semantic layer.

Prior to translation, the system scans the IR to extract domain-specific terminology and build a dynamic glossary. This extracted or user-provided glossary is then injected into the LLM prompt to improve terminological consistency across long documents. Furthermore, the IR enables the engine to logically stitch together paragraphs that are visually split across different columns or pages, reducing the semantic truncation common in naive PDF translators.

\subsection{Adaptive Typesetting Mechanism}

Due to differences in character sets and grammatical structures across languages, cross-lingual translation often leads to text expansion (e.g., English to Spanish). This expansion may cause the translated text to exceed the spatial constraints of the original layout.

To address this, our system incorporates an adaptive typesetting module that uses an iterative search mechanism to determine the minimum scaling factor required to accommodate the translated content for each paragraph. The process begins with an initial scaling factor of $1.0$ (representing no scaling). The system evaluates whether the translated text fits within the original paragraph bounding box defined in the IR. If the content overflows, the scaling factor is decremented by a predefined step size---typically $0.05$ or $0.10$---and the layout is recalculated. This iteration continues until the translated text is contained within the original boundaries or reaches a preset lower bound, yielding a localized scaling factor that balances readability and layout preservation.

\subsection{Nested Structure and CTM Reconstruction}

During PDF document translation, the parsing and restoration of nested structures are critical for layout stability. Commonly encountered elements in academic documents---including XObjects, Forms, path clipping, and composite structures---involve complex contextual dependencies and matrix transformations. Failure to address these may lead to text misalignment, formula anomalies, or clipping failures.

To address these issues, BabelDOC features a nested structure processing module. First, a context management unit maintains an XObject stack and a graphics state/clipping path stack, performing paired push/pop operations upon entering or exiting nested objects. This ensures the contextual independence of fonts, colors, and coordinate systems. Second, a paragraph composite expansion unit uniformly models character groups and inline images as paragraph sub-units, ensuring logical continuity. Finally, the rendering coordination unit applies graphics states layer-by-layer based on the Current Transformation Matrix (CTM), relocation transformations, and clipping paths. By scheduling the drawing of characters and curves according to rendering orders, the system flattens complex nested objects into stable drawing units, thereby improving visual reconstruction.

\section{Experiments and Evaluation}

\begin{table*}[t]
\centering
\resizebox{\textwidth}{!}{
\begin{tabular}{ll >{\columncolor{green!10}}c ccccc}
\toprule
\textbf{Category} & \textbf{Feature} & \textbf{BabelDOC} & \textbf{PDFMathTrans.} & \textbf{Doc2X} & \textbf{TeX-based} & \textbf{Google} & \textbf{DeepL} \\
\midrule
\multirow{2}{*}{\textbf{Accessibility}}
& Open-source Deployment           & \cmark & \cmark & \xmark  & \cmark & \xmark & \xmark \\
& API Integration                  & \cmark & \cmark & \cmark  & \cmark & \cmark & \cmark \\
\midrule
\multirow{7}{*}{\textbf{Layout \& Visual Integrity}}
& Multi-column Layout Preservation & \cmark & \cmark & \cmark  & \xmark & \xmark & Partial\\
& Vector Formula Translation       & \cmark & \cmark & \cmark  & \cmark & \xmark & \xmark \\
& Dual-page Bilingual Output       & \cmark & \cmark & Partial & \cmark & \xmark & \xmark \\
& Scanned PDF (OCR Support)        & \cmark & \cmark & \cmark  & \xmark & \cmark & \cmark \\
& Footnote Reconstruction          & \cmark & \xmark & Partial & \xmark & \xmark & \xmark \\
& Font Style Preservation          & \cmark & Partial& \xmark  & \cmark & \xmark & \xmark \\
& In-Figure/Table Text Translation & \cmark & \xmark & Partial & \xmark & \xmark & Partial\\
\midrule
\multirow{4}{*}{\textbf{Semantic \& NLP Control}}
& \textbf{Intermediate Rep. (IR)}  & \textbf{\cmark} & \textbf{\xmark} & \textbf{\xmark}  & \textbf{\xmark} & \textbf{\xmark} & \textbf{\xmark} \\
& Cross-page Context Stitching     & \cmark & \xmark & \xmark  & \xmark & \xmark & \xmark \\
& Auto Terminology Extraction      & \cmark & \xmark & \xmark  & \xmark & \xmark & \xmark \\
& Dictionary (Glossary) Constraints& \cmark & Partial& \xmark  & \xmark & \xmark & \cmark \\
\midrule
\multirow{2}{*}{\textbf{Efficiency}}
& Batch Processing                 & \cmark & \cmark & \cmark  & \cmark & \cmark & \cmark \\
& Avg. Speed (sec/page)            & 1.63  & 1.47   & 1.86    & 1.67   & 0.38   & 1.88   \\
\bottomrule
\end{tabular}
}
\caption{Comprehensive capability matrix of PDF document translation systems. PDFMathTrans. denotes PDFMathTranslate. TeX-based refers to workflows that reconstruct documents from available \LaTeX~source rather than directly translating arbitrary PDF inputs. The comparison reflects the evaluated configurations and publicly documented capabilities rather than all possible integrations of each system. Speed is reported for reference only, as systems differ in supported functions, rendering coverage, and deployment settings.}
\label{tab:full-tool-comparison}
\end{table*}

To evaluate BabelDOC, we conduct a multi-dimensional assessment focusing on layout preservation and semantic consistency across diverse document types. The evaluation characterizes BabelDOC's end-to-end behavior across representative document types, with emphasis on layout preservation, translation usability, and document-level consistency.

\subsection{Benchmark and Baselines}

\textbf{Dataset Selection.} We curated a benchmark of \textbf{200 complex PDF pages} to evaluate the system's robustness. The dataset spans three domains: (1) \textbf{Scientific Literature (80 pages)} from arXiv (Physics and Mathematics), characterized by dense formulas and multi-column layouts; (2) \textbf{Technical Documentation (60 pages)}, featuring hierarchical headers, nested lists, and code blocks; and (3) \textbf{International Patents (60 pages)}, containing high-density text and specialized terminology.

\textbf{Baselines.} We benchmark BabelDOC against:
\begin{itemize}
    \item \textbf{DeepL Document Translation:} A leading commercial document translation service for document-level translation.
    \item \textbf{PDFMathTranslate:} Our previous end-to-end precursor \cite{ouyang-etal-2025-pdfmathtranslate}, which preserves layout but does not expose an explicit IR layer.
\end{itemize}

\subsection{Evaluation Metrics}

We employ a hybrid evaluation strategy combining automated layout metrics and human-centric scoring:
\begin{itemize}
    \item \textbf{BIoU (BBox IoU):} Measures the geometric overlap between source and reconstructed layout elements, including paragraphs, tables, figures, and formula regions.
    \item \textbf{Subjective Scoring:} 200 samples are rated on a Likert scale (1--5) for \textbf{Layout Fidelity (LF)}, \textbf{Translation Precision (TP)}, \textbf{Visual Aesthetics (VA)}, and \textbf{Terminology Consistency (TC)}. We also report the average number of \textbf{Untranslated Text Blocks (UTB)} per page.
\end{itemize}

\paragraph{BBox IoU Calculation.}
BIoU is computed from layout-element bounding boxes extracted from each source page and translated page using the same parser. Coordinates are normalized by page size. Elements are matched by reading order and spatial proximity, and the final BIoU is averaged over matched elements and pages.

\begin{figure*}[t]
\centering
    \includegraphics[width=0.95\textwidth]{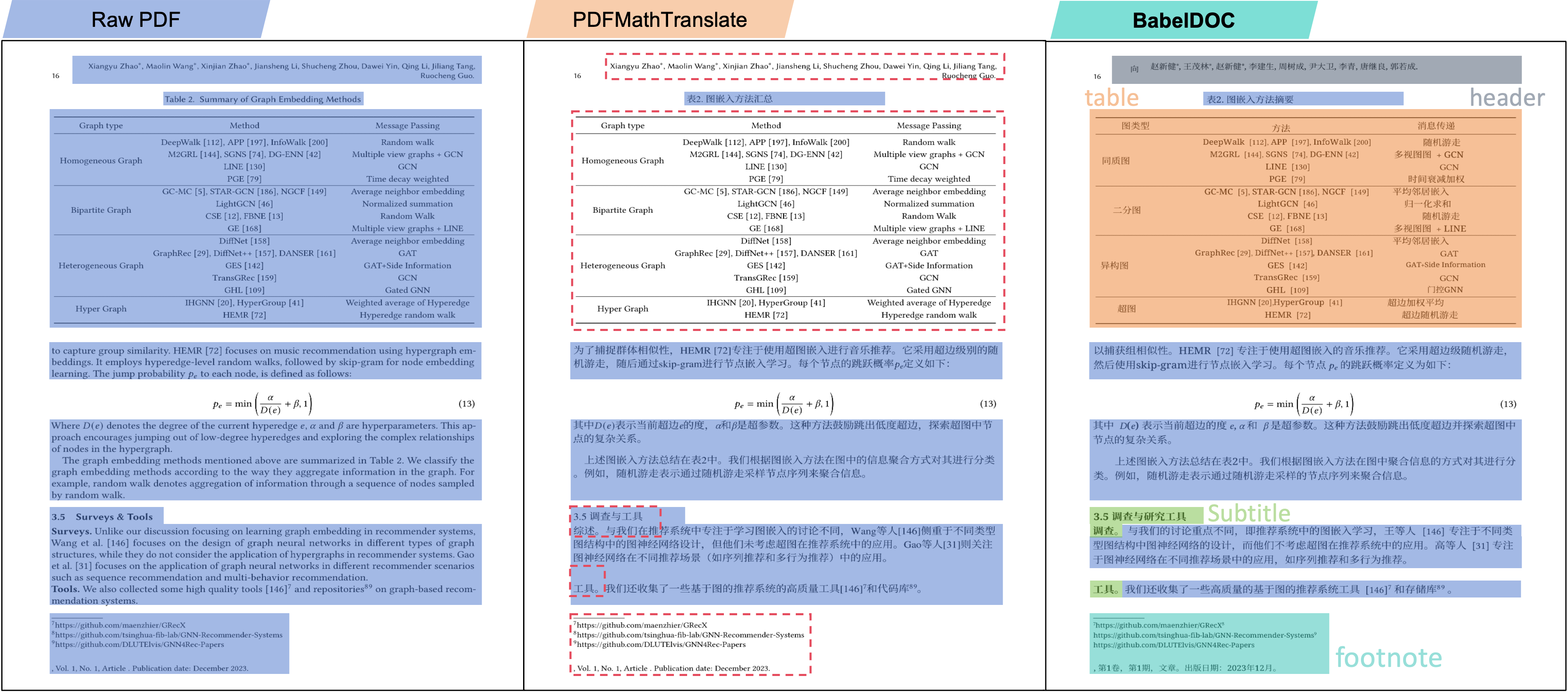}
    \caption{Qualitative results showing BabelDOC's capability in complex document translation. The method reconstructs multi-line equations and tables, preserves layout and fonts, and improves terminology consistency compared with the baseline method.}
    \label{fig:case_study}
\end{figure*}

\subsection{Human and LLM-as-a-Judge Evaluation}

To complement automated metrics, we implemented a dual-judge framework. First, three annotators with NLP backgrounds and bilingual proficiency in English and Chinese independently evaluated the translated pages. The system identities were anonymized, and the presentation order was randomized for each page. Each annotator scored layout fidelity, translation precision, visual aesthetics, and terminology consistency on a 1--5 Likert scale. We report the mean score across annotators and pages. We additionally counted untranslated text blocks as a page-level diagnostic metric.

Second, we utilized Gemini-2.5-Flash as a multi-modal judge. High-resolution images of the original and translated pages were provided to the LLM, which was prompted to act as a professional academic editor and score visual and semantic alignment. The LLM-as-a-Judge evaluation provides an additional multimodal assessment of visual and semantic alignment, complementing the human evaluation.

\begin{table}[h]
\centering
\resizebox{\columnwidth}{!}{%
\begin{tabular}{lcccccc}
\toprule
\textbf{System} & \textbf{BIoU}$\uparrow$ & \textbf{LF}$\uparrow$ & \textbf{TP}$\uparrow$ & \textbf{VA}$\uparrow$ & \textbf{TC}$\uparrow$ & \textbf{UTB}$\downarrow$ \\
\midrule
\multicolumn{7}{c}{\textit{Automatic Layout Metric}} \\
\midrule
DeepL (Doc)       & 19.8\% & - & - & - & - & - \\
PDFMathTrans.     & 48.7\% & - & - & - & - & - \\
\rowcolor{green!10}\textbf{BabelDOC} & \textbf{50.0\%} & - & - & - & - & - \\
\midrule
\multicolumn{7}{c}{\textit{LLM-as-a-Judge (Gemini-2.5-Flash)}} \\
\midrule
DeepL (Doc)       & - & 4.20 & \textbf{4.19} & 4.24 & 4.38 & 2.03 \\
PDFMathTrans.     & - & 2.55 & 2.78 & 2.54 & 3.02 & 5.55 \\
\rowcolor{green!10}\textbf{BabelDOC} & - & \textbf{4.46} & \textbf{4.19} & \textbf{4.49} & \textbf{4.43} & \textbf{1.70} \\
\midrule
\multicolumn{7}{c}{\textit{Human Evaluation}} \\
\midrule
DeepL (Doc)       & - & 3.44 & 3.62 & 3.63 & 4.21 & \textbf{2.33} \\
PDFMathTrans.     & - & 3.29 & 3.40 & 3.28 & 3.34 & 6.25 \\
\rowcolor{green!10}\textbf{BabelDOC} & - & \textbf{4.59} & \textbf{4.28} & \textbf{4.46} & \textbf{4.47} & 2.85 \\
\bottomrule
\end{tabular}%
}
\caption{Automatic layout metric and subjective scores on the 200-page benchmark. BIoU is computed from extracted layout-element bounding boxes. UTB denotes the average number of untranslated text blocks per page, where lower is better.}
\label{tab:subjective-eval-merged}
\end{table}

\subsection{Component Ablation}

To further analyze the contribution of key system components, we conduct an ablation study on 80 representative pages selected from the benchmark, including formula-heavy scientific pages, multi-column/table pages, and terminology-dense technical or patent pages. The ablated outputs are evaluated using the same 1--5 scoring scale as the subjective evaluation, focusing on layout fidelity, visual aesthetics, and terminology consistency. We select these dimensions because they are directly affected by adaptive typesetting and glossary/context control.

\begin{table}[t]
\centering
\resizebox{\columnwidth}{!}{%
\begin{tabular}{lccc}
\toprule
\textbf{Variant} & \textbf{LF}$\uparrow$ & \textbf{VA}$\uparrow$ & \textbf{TC}$\uparrow$ \\
\midrule
Full BabelDOC & 4.50 & 4.50 & 5.00 \\
w/o adaptive typesetting & 3.00 & 2.50 & 4.00 \\
w/o glossary/context control & 4.50 & 4.50 & 3.00 \\
\bottomrule
\end{tabular}%
}
\caption{Component ablation on representative pages. LF, VA, and TC denote layout fidelity, visual aesthetics, and terminology consistency, respectively.}
\label{tab:mini_ablation}
\end{table}

The ablation results show how these components affect different aspects of the system output. Adaptive typesetting mainly affects layout fidelity and visual aesthetics by reducing text overflow, while glossary/context control mainly affects terminology consistency in long technical documents.

\subsection{Results and Discussion}

As shown in Table \ref{tab:subjective-eval-merged}, BabelDOC achieves the strongest human scores for layout fidelity, translation precision, visual aesthetics, and terminology consistency among the evaluated systems. Its advantage is particularly clear in layout-related dimensions, where the IR-based reconstruction preserves multi-column structures, mathematical notation, and table regions more effectively than the baselines. The translation precision result is more nuanced: in the LLM-as-a-Judge evaluation, BabelDOC and DeepL obtain the same TP score, suggesting that BabelDOC's main benefit lies not in replacing strong translation engines, but in providing a layout-aware and controllable document-level translation pipeline around them.

Although BabelDOC improves layout fidelity, visual aesthetics, and terminology consistency, DeepL yields fewer untranslated blocks in the human evaluation. This suggests that some of BabelDOC's unresolved errors may stem from extraction-side omissions, OCR-side omissions, or rendering-stage coverage gaps. Overall, the results highlight BabelDOC's value as a high-fidelity document translation framework with layout-aware and controllable document-level processing.

\section{Qualitative Case Study}

To demonstrate the practical advantages of BabelDOC in handling complex scholarly documents, we present a qualitative comparison between the source PDF and its translated output, as illustrated in Figure \ref{fig:case_study}.

\subsection{Preservation of Structural Elements}

As shown in Figure \ref{fig:case_study}(a), OCR-based or text-only translators often struggle to render complex mathematical notation, either discarding it or producing misaligned text. By leveraging the \textit{Vector Formula Analyzer} derived from the IR, BabelDOC reconstructs vector-based formulas with high fidelity. The baseline offsets and script scales (superscripts/subscripts) are preserved, ensuring that the translated document remains useful for technical verification.

\subsection{Handling of Typographic Expansion}

A common failure mode in document translation is ``text overflow,'' where translated content (e.g., from English to German) exceeds the original bounding boxes. Figure \ref{fig:case_study}(b) highlights our \textit{Adaptive Typesetting Engine} in action. By dynamically calculating a localized scaling factor (in this case, $\gamma = 0.85$), BabelDOC fits the expanded translated text into the original multi-column grid while reducing overlap with adjacent figures or tables.

\subsection{Lexical Consistency across Multi-page Context}

One advantage of the IR-centric approach is document-level coherence. In this case study, a domain-specific term, \textit{``Current Transformation Matrix (CTM)''}, appears across non-contiguous pages. Naive translators may provide inconsistent variants, such as translating ``Matrix'' differently across paragraphs. As shown in Figure \ref{fig:case_study}(c), BabelDOC's pre-translation terminology extraction helps translate the specific term consistently throughout the document, facilitated by explicit IR-to-prompt glossary injection.

\section{Conclusion}

We presented BabelDOC, an open-source framework for high-fidelity document translation. By introducing a decoupled Intermediate Representation (IR), BabelDOC reduces the conflict between structural layout preservation and document-level semantic processing. The system empowers users with controllable NLP interventions---such as terminology constraints and cross-page contextual modeling---while improving visual integrity through adaptive typesetting. Evaluations on a curated 200-page benchmark show that BabelDOC improves layout fidelity, visual aesthetics, and terminology consistency over the evaluated baselines, while translation precision remains comparable to strong commercial document translation systems. Backed by an active open-source ecosystem, BabelDOC provides an accessible and extensible foundation for cross-lingual document intelligence. Future work will focus on real-time collaborative editing features, extraction robustness, and broader linguistic coverage.

\section*{Limitations}

Despite its practical utility, BabelDOC exhibits several constraints. First, the construction of the multi-modal Intermediate Representation (IR) introduces computational overhead. Compared to lightweight, text-only API calls, the deep layout analysis and adaptive typesetting phases increase inference latency, which may pose bottlenecks for real-time, high-concurrency synchronization on resource-constrained hardware. Second, the system's structural extraction relies on the efficacy of upstream OCR and layout detection models; heavily corrupted or highly unconventional scanned PDFs may still yield parsing anomalies that propagate to the typesetting engine. Third, BabelDOC may still encounter formatting errors on certain document types. In pages with dense overlapping objects, unusual clipping paths, or corrupted source encodings, the reconstruction module may produce local misalignment or untranslated regions. Our human evaluation also indicates that some commercial document translation systems may yield fewer untranslated text blocks on certain pages, suggesting that extraction robustness remains an important direction for future improvement. Finally, while the system supports dynamic scaling, translating between languages with extreme morphological and typographic divergences (e.g., transitioning from vertical traditional scripts to horizontal alphabetic scripts) remains an area for future systematic optimization.

\section*{Ethics Statement}

BabelDOC is released as an open-source engine under the GNU Affero General Public License v3.0 (AGPLv3) to foster reproducibility and support the research community in cross-lingual document intelligence. This copyleft licensing ensures that the core rendering engine remains open, supporting high-fidelity document translation for the global open-science community. This paper focuses on the underlying layout-aware methodology, the intermediate representation framework, and the architectural advancements. Users should be aware that translation systems may introduce errors or omissions, especially in technical, legal, or medical documents, and outputs should be reviewed by qualified professionals before high-stakes use.

\section*{Acknowledgments}

This work was supported by the National Natural Science Foundation of China under Grant No. 62306173. We extend our gratitude to the open-source NLP and computer vision communities. Specifically, we acknowledge the foundational contributions of the Doc2X, MinerU, and Surya projects, which inspired our layout parsing methodologies. We also thank the contributors and users on GitHub whose testing and feedback have been instrumental in refining the BabelDOC ecosystem.

\bibliography{custom}

\appendix
\label{sec:appendix}

\section{Intermediate Representation (IR) Data Structure}
\label{sec:ir_structure}

To elucidate the mechanism of our decoupled translation paradigm, we provide a simplified JSON snippet of the Intermediate Representation (IR). The IR standardizes raw PDF imperatives into a structured, bidirectional hierarchical schema. As shown in Listing \ref{lst:ir_json}, the IR isolates visual metadata (e.g., \texttt{pdf\_font}, \texttt{page\_layout}, and exact \texttt{bounding boxes}) from the semantic \texttt{paragraph} text. The \texttt{placeholders} array within the translation tracking block demonstrates how non-translatable structured fragments, such as citation markers, formula spans, and vector-rendered symbols, can be masked before LLM intervention, reducing the risk of placeholder corruption during translation.

\begin{lstlisting}[caption={A simplified snippet of BabelDOC's Intermediate Representation (IR), illustrating the decoupling of layout metadata, character-level graphic states, and semantic NLP translation tracking with structured placeholders.}, label={lst:ir_json}, float=hbt!]
{
  "page_number": 0,
  "unit": "point",
  "page_layout": [
    {
      "id": 1,
      "class_name": "page_header_hybrid",
      "box": {"x": 79, "y": 653, "x2": 441, "y2": 663},
      "conf": 1.0
    }
  ],
  "pdf_font": [
    {
      "font_id": "F189",
      "name": "KYQJPC+LinBiolinumTB",
      "ascent": 693,
      "descent": -234
    }
  ],
  "pdf_character": [
    {
      "char_unicode": "1",
      "font_size": 7.97,
      "box": {"x": 45.6, "y": 646.5, "x2": 49.4, "y2": 654.5},
      "render_order": 1
    }
  ],
  "paragraph": [
    {
      "input": "DeepWalk {v1}APP {v2}InfoWalk{v3}",
      "output": "DeepWalk {v1}APP {v2}InfoWalk{v3}",
      "pdf_unicode": "DeepWalk [112], APP [197], InfoWalk [200]",
      "layout_label": "table_cell_hybrid",
      "placeholders": [
        {
          "type": "citation_marker",
          "id": 1,
          "placeholder": "{v1}",
          "source_chars": "[112],"
        }
      ]
    }
  ]
}
\end{lstlisting}

\section{Usage and Accessibility}

While BabelDOC serves as a standalone intermediate representation backend, we have engineered a user-oriented ecosystem to maximize its accessibility across both technical and non-technical users.

To bridge the gap between algorithmic research and practical deployment, we provide programmable Python APIs accessible via the \texttt{pdf2zh\_next} module for integration into larger document intelligence pipelines, alongside Dockerized instances for secure, isolated local deployment.

For end-users, the framework provides an intuitive Command Line Interface (CLI) deployable via PyPI (e.g., \texttt{pip install BabelDOC} or via \texttt{uv}), a Windows-native Graphical User Interface (GUI), and a web-based interactive studio. By executing the \texttt{babeldoc} command, users can leverage the underlying IR framework to inject custom translation glossaries via CSV, configure scaling constraints, and utilize a dual-page alternating layout mode (\texttt{--use-alternating-pages-dual}) to generate bilingual reading materials.

\section{Sustainability and Extensibility}

To ensure the long-term sustainability of BabelDOC, our development strategy centers on a modular, plugin-based architecture. The explicit decoupling of textual content from layout information provides a stable API boundary, allowing researchers to hot-swap emerging Vision-Language Models (VLMs) for parsing or state-of-the-art LLMs, e.g., GPT, Claude, Qwen, and GLM, for generation, without altering the core rendering engine.

Furthermore, as an open-source initiative, BabelDOC facilitates collaborative maintenance and rapid community-driven iteration. The project has attracted over 8.4K GitHub stars and 17 contributors at the time of writing, indicating active community interest and continued open-source maintenance.

\section{LLM-as-a-Judge Evaluation Prompt}
\label{sec:judge_prompt}

To ensure a standardized evaluation, we utilized Gemini-2.5-Flash as a multimodal judge. The model was provided with high-resolution images of the original PDF and the results from the three translation systems. The specific system prompt used for the comparative scoring is provided below:

\begin{quote}
\small
\textbf{Role:} You are a senior Academic Journal Editor performing a comparative evaluation of PDF translations produced by three different translation systems.

\textbf{Task:} Compare ALL THREE translated pages against the Original simultaneously. Because you are comparing them side-by-side, score them \textbf{relative to each other}.

\textbf{Evaluation Rubrics (1-5 each):}
\begin{itemize}
    \item \textbf{Layout Fidelity:} Maintenance of columns, margins, font hierarchies, positioning of figures/tables.
    \item \textbf{Translation Precision:} Accuracy of academic meaning, scientific claims, data descriptions.
    \item \textbf{Visual Aesthetics:} Professional typography, line spacing, absence of text overlaps or bleeding.
    \item \textbf{Terminology Consistency:} Uniform use of domain-specific jargon, citations, figure labels.
\end{itemize}

\textbf{Untranslated Blocks Count:} Count the number of distinct text blocks that remain in the original language. Any text block that appears identical to the Original (still in English) should be counted. Output the integer count.

\textbf{Output Format:}
\texttt{system|Layout Fidelity:<score>|Translation Precision:<score>|Visual Aesthetics:<score>|Terminology Consistency:<score>|Untranslated Blocks:<count>}
\end{quote}

\end{document}